\documentclass[11pt]{article}

\usepackage[final]{acl}

\usepackage{times}
\usepackage{latexsym}

\usepackage[T1]{fontenc}
\usepackage[utf8]{inputenc}

\usepackage{microtype}

\usepackage{inconsolata}

\usepackage{graphicx}

\title{Syntactically-guided Information Maintenance in Sentence Comprehension}

\author{Shinnosuke Isono \\
  NINJAL \\
  \texttt{s-isono@ninjal.ac.jp} \\\And
  Kohei Kajikawa\\
  Georgetown University \\
  \texttt{kk1571@georgetown.edu} \\}

\makeatletter\def\new@fontshape{}\makeatother
\usepackage{linguex}

\begin{document}
\setlength{\Exlabelsep}{0.5em}
\maketitle

\begin{abstract}
Maintaining information in context is essential in successful real-time language comprehension, but maintenance is cognitively costly and can slow processing. We hypothesize that rational language users selectively maintain information that is crucial for future prediction, guided by syntactic structure. Under this view, two factors affect maintenance cost: the number of predicted heads and the number of incomplete dependencies. Although these factors have been treated as competing hypotheses in the literature, our account predicts that they are not reducible to one another. We show this is the case in a naturalistic reading time dataset in Japanese, a language in which the two factors contrast particularly clearly. We further show that there is a tradeoff such that readers that slow down for maintenance tend to benefit more from predictability, providing additional support for the proposed account. These patterns are not evident in English, however, and we highlight some issues to be resolved to understand the contribution of syntax in memory-efficient processing of various languages.
\end{abstract}

\section{Introduction}
``Language is fleeting,'' as \citet{christiansen-chater-2016-now} put it. Language unfolds over time, and comprehension takes place incrementally \cite{tanenhaus-etal-1995-integration}, but a human language user cannot keep track of everything that is said or written. Yet words are related to each other, often at a distance, and the language user needs to work out their relations to fully understand the meaning of the expression. From a different viewpoint, the language user should seek to minimize the error in prediction using the past input as the context \cite{hale-2001-probabilistic,levy-2008-expectation}. How do humans manage to do that with their limited memory capacity?

We hypothesize that a rational language user saves memory resources by selectively maintaining words that are critical for predicting future inputs \cite{hahn-etal-2022-resource, xu-futrell-2026}, and syntactic structure provides a reliable clue for such maintenance by limiting words that can be directly related to upcoming words (Figure \ref{fig:illustration}).

\begin{figure}[t]
  \includegraphics[width=\columnwidth]{img/illustration.png}
  \caption{Incomplete dependencies and predicted heads in example sentence fragments in English (a verb-medial language) and Japanese (a verb-final language).}
  \label{fig:illustration}
\end{figure}

This view revives an old question about \textbf{maintenance (storage) cost}. Maintaining linguistic information should be cognitively costly \cite{yngve-1960-model, miller-chomsky-1963-finitary, kimball-1973-seven, gibson-1991-computational, stabler-1994-finite, lewis-1996-interference, gibson-1998-linguistic, gibson-2000-dependency}, and can be observed as reading slowdown. In terms of dependency structure, the number of \textbf{predicted heads}\footnote{Here, \emph{heads} refers to atoms of dependency analysis. We do not mean to contrast heads with dependents. If one element triggers prediction of another element, that counts as a predicted head, regardless of the direction of the dependency.} is known to be a major determinant of maintenance cost crosslinguistically \cite{chen-etal-2005-online, nakatani-gibson-2010-online, ristic-etal-2022-maintenance, isono-etal-2025-modeling}. In the current account, this is because heads being predicted means some bits of information that are valuable for future prediction are carried.\footnote{See also \citet{kajikawa-etal-2026} for an information-theoretic formulation.} Importantly, the current account also predicts that additional \textbf{incomplete dependencies} to be completed at the same predicted head further increase the maintenance cost \cite[cf.][]{gibson-1991-computational}, since they carry additional information. Previous studies, in contrast, treat predicted heads and incomplete dependencies as competing explanations for the maintenance cost, and empirical evidence has been taken to favor predicted heads \cite{gibson-1998-linguistic,nakatani-gibson-2008-distinguishing}.

We reinvestigate the contribution of predicted heads and incomplete dependencies to the maintenance cost during reading, using a large-scale naturalistic reading time dataset in Japanese. Japanese is suitable for the current purpose because these two factors contrast most clearly in a strictly head-final language. The use of large-scale dataset and the statistical control for predictability may reveal the maintenance cost of incomplete dependencies which could have been masked in the previous study by being too small or because of a confounding factor (see Section \ref{sec:related}). We indeed find that both predicted heads and incomplete dependencies have a slowdown effect that is not reducible to the other, with the former being larger, as predicted by the current account.

There is another ``rational'' strategy that a reader can take. When there are too many pieces of information, one can give up maintaining everything and hasten to get out of the demanding structure \cite[cf.][]{ferreira-etal-2002-good}. The drawback is that the future input is less predictable since the context is forgotten. We show that such a tradeoff exists, adding further support for the view that maintenance is for better prediction.

While a strictly head-final language like Japanese is the most suitable to address the current question, it is important to see whether the pattern generalizes to typologically different languages, as reviewers pointed out. We additionally conducted the same analysis with an English dataset. The number of predictive heads turns out to be a significant predictor, as in Japanese, when content words are analyzed. Beyond that, however, we do not find evidence for the patterns observed in Japanese. There are some issues in implementing a syntactically-guided memory-efficient predictive processing in typologically diverse languages, and we discuss them in Limitations.

The rest of this paper is structured as follows. Section \ref{sec:background} introduces the background and describes the current hypothesis in more detail. Section \ref{sec:experiment-1} reports the experiment evaluating the two metrics of maintenance cost in Japanese. Section \ref{sec:experiment-2} reports the follow-up experiment investigating the tradeoff between maintenance and prediction. Section \ref{sec:experiment-3} reports the experiment in English. Section \ref{sec:discussion} discusses the implications of the results, related work, and limitations. The code is available online.\footnote{\url{https://osf.io/cvket/overview?view_only=0b8d2db223274da981f7ceea1d39b5f5}}

\section{Background} \label{sec:background}
\subsection{Syntax as a clue for rational maintenance}
A key observation that motivates the current study is that a subset of words in the context is highly informative in predicting the upcoming words. Consider predicting the next word following \Next.

\ex. The professor who supervises Alex will \label{ex:english}

\textit{Will} suggests that the next word is likely to be an infinitival verb like \textit{be} and \textit{teach}, rather than a finite verb, noun, adjective, etc. \textit{Professor} suggests that the next word is likely to be an action that a professor often does, such as \textit{teach} or \textit{publish}, and is less likely to be actions like \textit{cook} and \textit{roar}. In contrast, other words are less informative: \textit{the, who, supervises,} and \textit{Alex} tell relatively little about what the next word is likely to be. The syntactic structure explains why. An analysis based on the Universal Dependency--style dependency grammar \cite{demarneffe-etal-2021-universal} of \Last is shown in Figure \ref{fig:illustration}. \textit{Professor} and \textit{will} are expected to depend on the same upcoming verb. Other words in the fragment are unlikely to form a dependency with upcoming words, as they have already found their head and there is a general ban on crossing dependencies. More generally, syntactic dependency structure tells which words in context are highly informative for future prediction. In fact, this is the source of the anti-locality effect documented in many languages \cite{konieczny-2000-locality,konieczny-doring-2003-anticipation,vasishth-lewis-2006-argument, asahara-etal-2019-bccwj, isono-2024-category}, where preceding dependents facilitates the processing of the head. Though other words can be informative through semantics and world knowledge in some contexts, the contribution of syntactically related elements is more general and stable.

This observation directly suggests a rational strategy that a memory-limited language processor that seeks to minimize prediction error would employ: maintaining \textit{professor} and \textit{will} while neglecting others. More generally, we hypothesize that the human language processor rationally allocates memory resources to words that are particularly informative for future prediction \cite[cf.][]{hahn-etal-2022-resource, xu-futrell-2026}, and syntax helps identify which words are. Words that are not maintained will be eventually forgotten by decay and/or interference \cite{lewis-vasishth-2005-activation}, unless there are other reasons to maintain them.

\subsection{Syntactic metrics of maintenance cost}
The syntax-based approach to resource-rational processing is also interesting since it brings a new perspective to an old question about the maintenance cost during processing. As mentioned in the Introduction, there are two possible ways to quantify maintenance cost under a dependency-style syntactic analysis: the number of predicted heads and the number of incomplete dependencies \cite{gibson-1998-linguistic}. While these two are correlated since a predicted head implies an incomplete dependency, they often diverge in head-final structures.

Japanese, for example, is a strictly head-final language. As a result, multiple dependents often precede the clause-heading verb. It has been shown that speakers can predict the clause structure before the head verb using case markers on those dependents \cite{kamide-mitchell-1999-incremental,miyamoto-2002-case}. In \Next, the markers suggest that the three nouns can all depend on a single ditransitive verb.

\exg. \label{ex:gaoni}Kyoju-ga gakusei-ni hon-o / \textcolor{red}{miseta}.\\
prof-\textsc{nom} student-\textsc{dat} book-\textsc{acc} {} showed\\
\glt `The prof. showed the student the book.'

At the point before the verb, indicated by the slash, \textbf{one} head is predicted (\textcolor{red}{the ditransitive verb}), while \textbf{three} dependencies remain incomplete. Compare \Last with \Next, where all the three nouns are marked as nominative:

\exg.\# Kyoju-ga gakusei-ga hon-ga / omosiroi-to omotteiru-to itta. \label{ex:gagaga}\\
Prof-\textsc{nom} student-\textsc{nom} book-\textsc{nom} {} interesting-that think-that said\\
\glt `The professor said that the student thought that the book was fun.'

At the slash, \textbf{three} heads are predicted and \textbf{three} dependencies remain incomplete. \citet{gibson-1998-linguistic} takes the apparent difficulty of \Last compared to \LLast as evidence that the number of predicted heads is a better metric of maintenance cost than the number of incomplete dependencies. Controlled experiments in English \citep{chen-etal-2005-online} and Japanese \citep{nakatani-gibson-2008-distinguishing} corroborate this view (we revisit the Japanese experiment in Section \ref{sec:related}). The effect of predicted heads is also observed in a Japanese naturalistic reading time dataset \cite{isono-etal-2025-modeling}, although that study does not compare predicted heads with incomplete dependencies.

Unlike previous work, the current resource-rational view sees predicted heads and incomplete dependencies as complementary, rather than competing, explanations of maintenance cost. For example, the dative-marked \textit{gakusei-ni} in \ref{ex:gaoni} and the nominative-marked \textit{gakusei-ga} in \ref{ex:gagaga} are both informative for future prediction, just differently. The former narrows down the type of the already predicted verb, while the latter implies another verb. The rational processor should maintain both types of information. Accordingly, both should affect the maintenance cost. The relative difficulty of \Last suggests that the latter type of information (predicted heads) is ``heavier'' than the former (incomplete dependencies).

\section{Experiment 1} \label{sec:experiment-1}
Our hypothesis predicts that the maintenance cost is affected by both the number of predicted heads and the number of incomplete dependencies, and the former has a greater impact on reading times. We test these predictions using a large-scale reading time dataset in Japanese.

\subsection{Materials}
We use the BCCWJ-SPR2 dataset \cite{asahara-2022-reading}. It is currently the largest dataset of naturalistic reading times in Japanese to our knowledge. It is based on the BCCWJ corpus \cite{maekawa-etal-2014-balanced} and contains self-paced reading times \cite{just-etal-1982-paradigms} from native speakers. Texts were presented one \emph{bunsetsu} at a time, where bunsetsu is a word-like unit commonly used in Japanese linguistics. A bunsetsu often contains multiple morphemes, e.g., a noun and a case marker, or a verb and a series of auxiliaries. For readability, bunsetsu will be referred to as region in the rest of this paper. We use the PB subset, which is a collection of excerpts from various books, since Universal Dependencies (UD) annotation is available for this subset.

For the current analysis, we take the mean reading time for each region in the dataset, aggregating the actual reading time from individual participants, and use it as the dependent variable \cite[e.g.,][]{goodkind-bicknell-2018-predictive, wilcox-etal-2023-testing}. As in many reading studies using naturalistic data, the sentence-final regions are excluded from analysis since they often reflect wrap-up effects. The first two regions of each sentence are also excluded since they can be affected by spillover from the final region of the previous sentence. After the exclusion, the data consists of 83 documents, 8,050 sentences, or 50,285 regions, coming from 424 participants.

This dataset is already shown to exhibit the effect of maintenance cost quantified as the number of predicted heads \cite{isono-etal-2025-modeling}. Our focus is therefore on the effect of incomplete dependencies and its relation to the effect of predicted heads.

\subsection{Variables}
\paragraph{Maintenance cost}
The variables of interest for the current study are the number of predicted heads and the number of incomplete dependencies at each region. They are computed from the 
UD--style syntactic annotation of the target dataset provided by \citet{asahara-etal-2018-universal}. We only consider these gold parses, while in reality the incremental parser may also pursue different parses (the ``perfect oracle'' assumption in the sense of \citet{brennan-2016-naturalistic}). The number of predicted heads is operationalized as the number of words that have a dependency with a word in the context but are yet to be seen. The number of incomplete dependencies is the number of dependencies that start with a word in the context and end with a word that is yet to be seen. These values are initially calculated at the morpheme level, following the UD annotation, and then aggregated at the region level by taking the minimal value. Taking the minimum virtually means that we only count dependencies that neither begin nor end in the region in question. This is necessary because the amount of region-internal dependencies would be strongly affected by the length of the region, and by the idiosyncrasy of UD annotation which attaches all the verbal suffixes to the stem.

\paragraph{Control variables}
The baseline model contains the following five control variables: the position of the sentence in the document, the position of the region in the sentence, the number of characters in the region, the sum of the unigram surprisal of the morphemes comprising the region, and the sum of the GPT-2 surprisal of the morphemes. We take surprisal \cite{hale-2001-probabilistic, levy-2008-expectation} into account not only because it is known to strongly affect reading times \cite{wilcox-etal-2023-testing,xu-etal-2023-linearity}, but because the confounding effect of expectation may be one reason that the cost from incomplete dependencies were not observed previously (see Section \ref{sec:related}). The unigram surprisal is calculated using the frequency table of the NINJAL Web Japanese Corpus \cite{asahara-etal-2014-archiving}. We obtain by-morpheme surprisal values and then summed them at the region level rather than directly assessing the by-region surprisal because the exact combination of morphemes can be hard to find in a corpus. GPT-2 surprisal, based on the GPT-2 large language model \cite{radford-etal-2019-language}, is calculated using the \texttt{japanese-gpt2-xsmall} model \cite{sawada-etal-2024-release}.\footnote{\url{https://huggingface.co/rinna/japanese-gpt2-xsmall}. The perplexity of the model for the current dataset is 61.1.} We use the \texttt{xsmall} model since larger models are known to fit worse to reading times both generally \cite{oh-schuler-2023-surprisal} and with the current dataset \cite{isono-etal-2025-modeling}.

\paragraph{Spillover}
In the self-paced reading paradigm, the processing load of one region often affects reading times of the subsequent regions. This is called the spillover effect \cite{mitchell-1984-evaluation}. Spillover effects are taken into account in the current analysis by including variables from $(i-1)$th and $(i-2)$th regions in the regression model for the $i$th region. We do this for the number of characters, unigram surprisal, and GPT-2 surprisal. Spillover of the position effects is not considered because the position of the previous words is perfectly predictable from the current position. Spillover of the maintenance cost is not considered either, because the maintenance cost of adjacent regions should correlate strongly, and because when they differ, it is because some dependencies have initiated or terminated there, but we do not want effects of dependency initiation or termination to be mixed with the maintenance cost.

\begin{figure}[t]
  \includegraphics[width=\columnwidth]{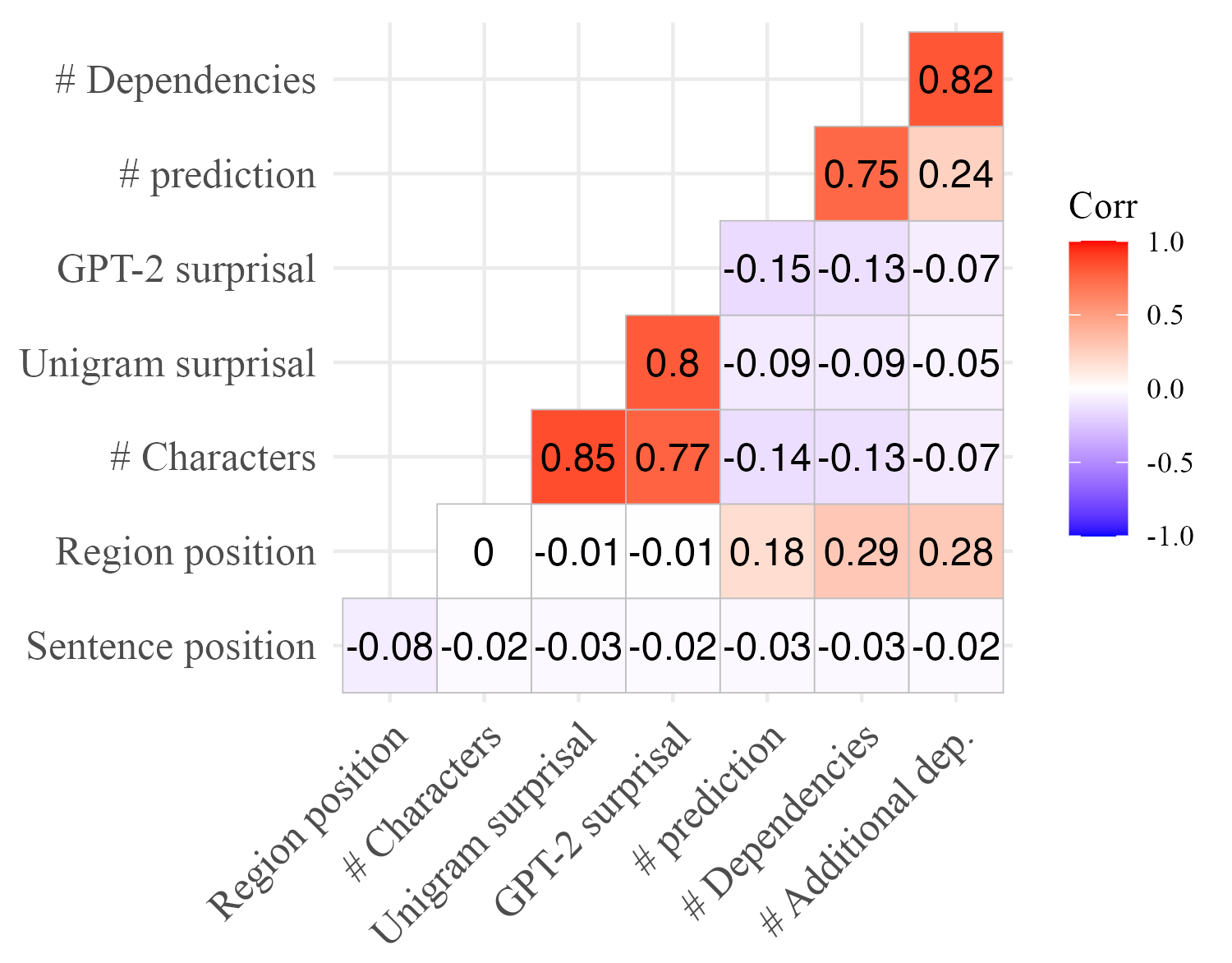}
  \caption{Correlations between variables used in the regression analysis. \# = number of.}
  \label{fig:correlations}
\end{figure}

\paragraph{Correlations}
Correlations between the variables are shown in Figure \ref{fig:correlations}. The two metrics of maintenance cost are highly correlated. This is because a predicted head by definition implies an incomplete dependency. This correlation can distort the estimate of the coefficients. We therefore use the number of \emph{additional} incomplete dependencies, which is simply the number of incomplete dependencies minus the number of predicted heads, when estimating the coefficients (Section \ref{sec:results.coefficients}).

\subsection{Analysis}
\paragraph{Psychometric predictive power}
Following studies on naturalistic reading time datasets \cite[among others]{goodkind-bicknell-2018-predictive, wilcox-etal-2023-testing}, we evaluate the psychometric predictive power of a maintenance cost variable by comparing linear regression models with and without it. Both models contain the control variables defined above. If the inclusion of a variable improves the model fit, that variable is considered as having psychometric predictive power. Specifically, we measure the decrease in the mean squared error ($\Delta$MSE) of the linear regression model by the inclusion of the maintenance cost variable. The error for each region is calculated by 10-fold cross-validation repeated 50 times, and a permutation test is conducted to test whether $\Delta$MSE is significantly above zero.

\paragraph{Coefficients}
We also check the regression coefficients assigned to the maintenance cost variables. If the psychometric predictive power of these variables reflects maintenance cost, the coefficients should be positive (i.e., slowdown effects). We also predict the coefficient for predicted heads to be larger (recall the comparison between \ref{ex:gaoni} and \ref{ex:gagaga}). For this analysis, we use the number of additional incomplete dependencies instead of the number of all incomplete dependencies. To obtain a reliable estimate of the effect, the data is randomly split into 10 folds and linear regression models are fit to each of them. This gives a distribution of estimated coefficients for each variable. We conduct permutation tests on these distributions to address the above questions.

\paragraph{Exclusion of potential hidden dependencies}
One potential concern for the current approach is that there are hidden dependencies that are not available in the UD annotation. Certain expressions in Japanese trigger prediction for a particular type of auxiliary or particle to be attached to the verb, but such relations are not expressed in the UD annotation. In the following example, \textit{-sika} `only' requires a negative auxiliary to be attached to the verb. In the UD annotation, however, \textit{-sika} attaches to the noun, and the noun depends on the head verb as usual; the dependency between \textit{-sika} and negation is not indicated.

\exg.
Kyoju-ga hon-\textbf{sika} kawa-\textbf{nakat}-ta.\\
Prof-\textsc{nom} book-only buy-not-ed\\
\glt `The professor only bought a book.'

There are many other negation-triggering expressions in Japanese. A similar problem arises with the UD analysis of a \textit{wh}-expression (e.g., \textit{dare} `who') and an obligatory question particle attached to the verb (usually \textit{ka}). Psycholinguistic evidence shows online processing is sensitive to such correspondences \cite{ono-nakatani-2014-integration, nakatani-2023-locality}. Such correspondences can result in a superficial effect of incomplete dependencies above and beyond predicted heads, because expressions like \textit{hon-sika} `book-only' in \Last increase the number of incomplete dependencies but not the number of predicted heads. For this reason, we also run an analysis using only sentences without any negation or question particles.\footnotemark{} This excluded 1,908 sentences (23.8\%).

\footnotetext{Particles are identified using the lemma column of the UD annotation.}

\subsection{Results}
\paragraph{Psychometric predictive power}
\begin{figure}[t]
  \includegraphics[width=\columnwidth]{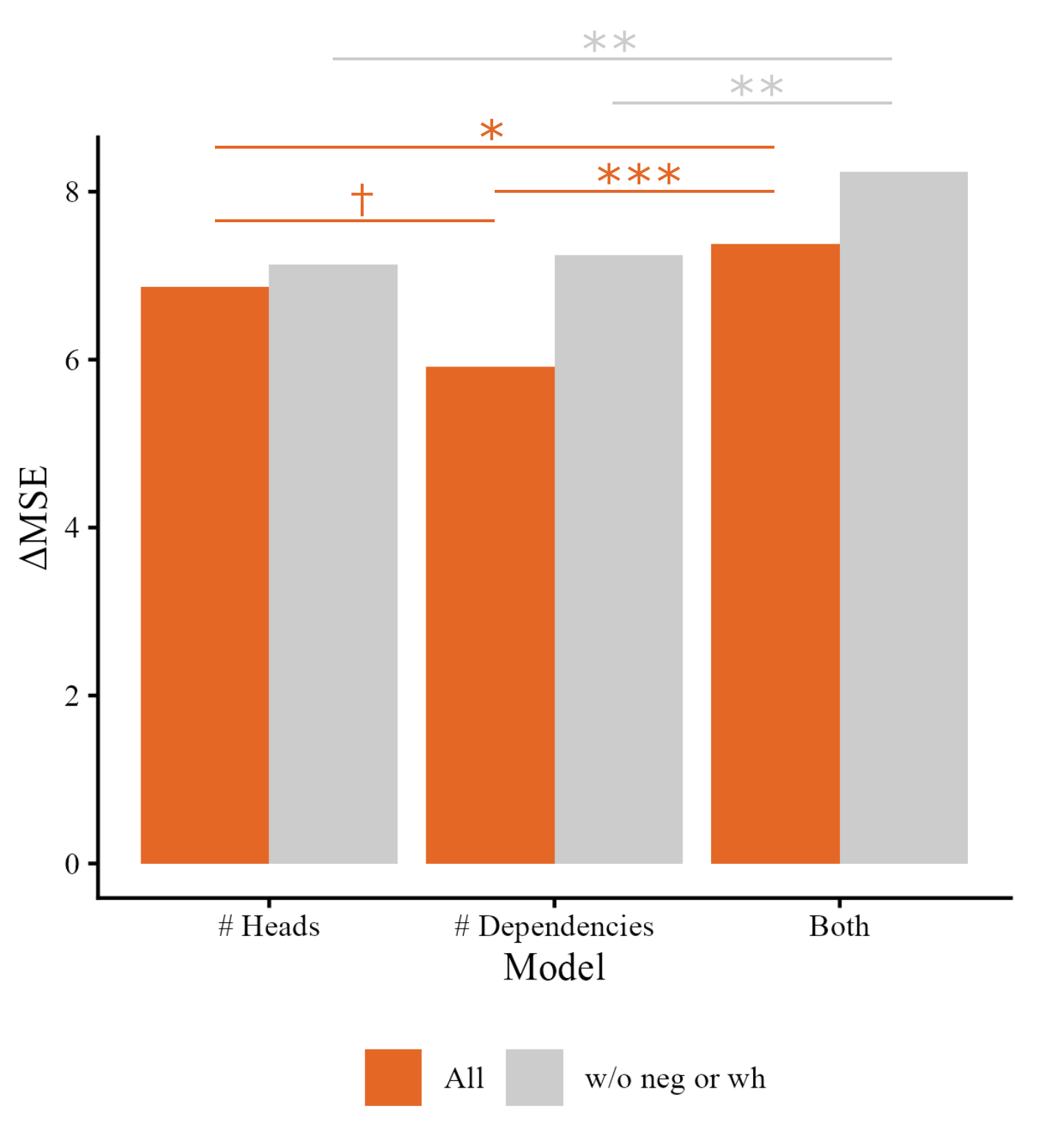}
  \caption{$\Delta$MSE per word by model. The stars indicate the significance of permutation tests: *** for $p<0.001$, ** for $p<0.01$, * for $p<0.05$, $\dagger$ for $p<0.1$}
  \label{fig:dmses}
\end{figure}

\begin{table}[t]
    \centering
    \begin{tabular}{lr}
         & $\Delta$MSE \\
         \hline
         Sentence position & 176.3\\
         Region position & 6.6\\
         \# Characters & 3.4\\
         Unigram surprisal & 331.9\\
         GPT-2 surprisal & 2.4\\
         \hline
         \# Heads & 6.9\\
         \# Dependencies & 5.9\\
         Both & 7.4\\
         \hline
    \end{tabular}
    \caption{$\Delta$MSE of predictors.}\label{tab:dmses}
\end{table}

We first compare the number of predicted heads and the number of incomplete dependencies by their psychometric predictive power. The results are summarized in Figure \ref{fig:dmses} (the orange bars). Both the number of predicted heads and the number of incomplete dependencies show a significantly positive $\Delta$MSE, meaning that both are predictive of human reading times. Predicted heads have a higher $\Delta$MSE than incomplete dependencies, though the difference is only marginally significant ($p=0.058$). The model that contains both predicted heads and incomplete dependencies performs significantly better than the models that contain only one of them ($p=0.023$ when compared against predicted heads, and $p<0.001$ when compared against incomplete dependencies). These results indicate that both predicted heads and incomplete dependencies have psychometric predictive power that cannot be reduced to that of the other.

The magnitude of the psychometric predictive power of the maintenance cost predictors is compared with baseline predictors in Table \ref{tab:dmses}. The $\Delta$MSE values for baseline predictors are calculated by comparing baseline models with and without that predictor. While the power is much smaller than those of some low-level predictors, it is larger than GPT-2 surprisal, which is supposed to capture high-level linguistic structure.

\paragraph{Coefficients} \label{sec:results.coefficients}
\begin{figure}[t]
  \includegraphics[width=\columnwidth]{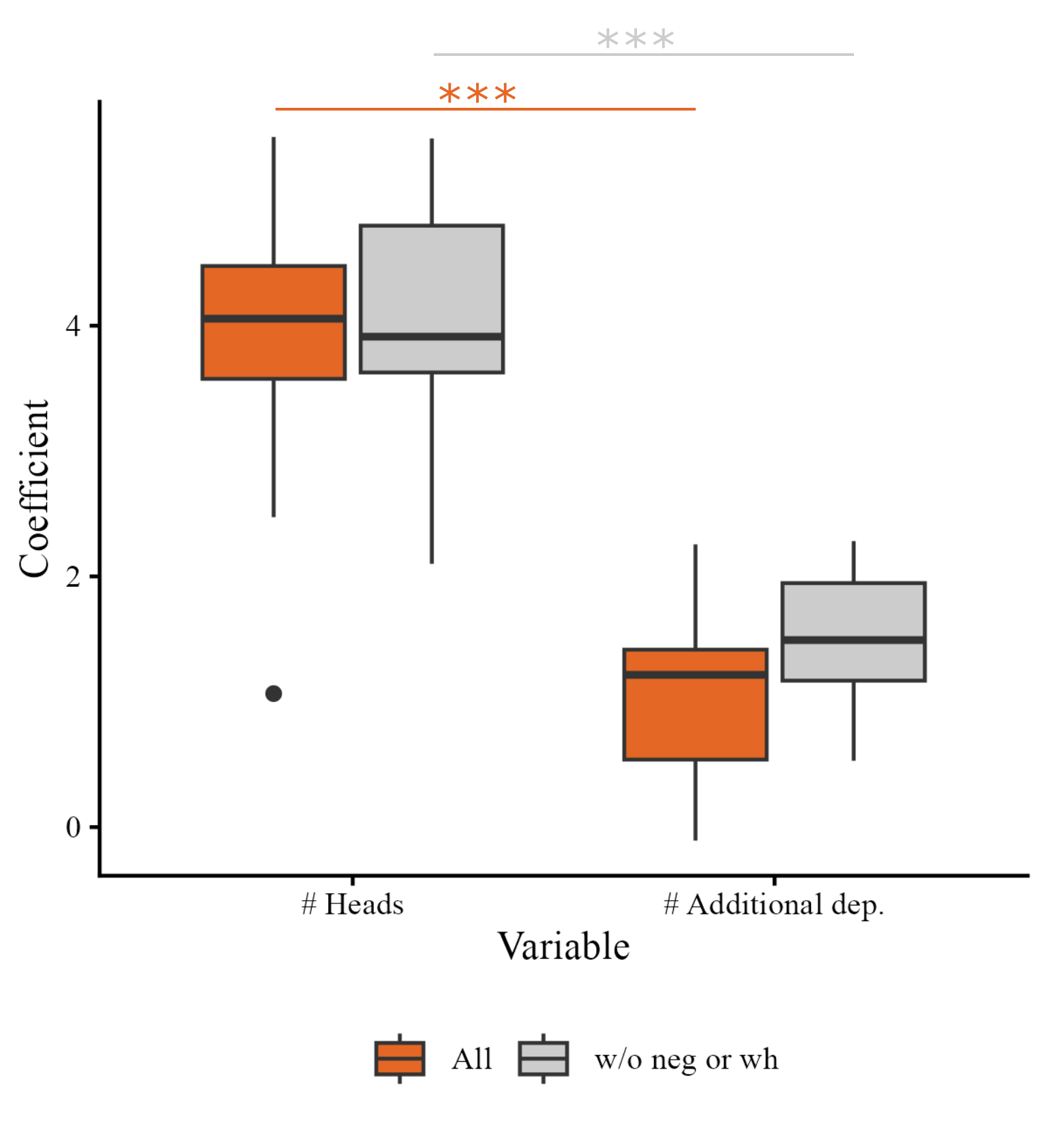}
  \caption{Distribution of estimated coefficients of the metrics of maintenance cost. The stars indicate the significance of permutation tests: *** for $p<0.001$}
  \label{fig:coefficients}
\end{figure}

The distribution of estimated coefficients for the maintenance cost variables are summarized in Figure \ref{fig:coefficients}. As can be seen, both variables show a consistent positive effect across the folds. The coefficient for the predicted heads is significantly larger than the coefficient for the incomplete dependencies ($p<0.001$). These results are consistent with the view that both predicted heads and incomplete dependencies impose maintenance cost, but the former is heavier.

\paragraph{Exclusion of potential hidden dependencies}
The above analyses are repeated with the subset of the original data that do not contain any negation or question particle. The results after exclusion are shown as gray bars in Figures \ref{fig:dmses} and \ref{fig:coefficients}. The exclusion does not change the key results: both of the two maintenance cost variables have psychometric predictive power, which cannot be reduced to that of the other, and the coefficient is larger for the number of predicted heads.

\section{Experiment 2}\label{sec:experiment-2}
So far we have analyzed the general tendency among readers. However, different readers can take different strategies of maintenance. Recall that a high maintenance cost is symptomatic of a center-embedding structure, like \ref{ex:gagaga}. Given the difficulty of such structures, some readers may give up mid-sentence maintaining the required information; then the maintenance cost would disappear. Some may even speed up in order to get out of the center-embedding structure, to at least comprehend the message carried by the matrix structure. These can be seen as instances of good-enough processing \cite{ferreira-etal-2002-good}. By giving up maintenance, readers are relieved from memory load, but they are sacrificing accurate parsing and prediction.

Below, we show such a tradeoff exists. First, we show that participants respond differently to the need of maintenance: many slow down but some speed up. Second, we show that this difference in maintenance strategy affects how well they predict.

\subsection{Materials}
We use the same BCCWJ-SPR2 dataset as in Experiment 1. Unlike in Experiment 1, we do not collapse data across participants. Instead, we model raw reading times from participants with 1,000 or more data points. 353 participants are included in this analysis, with 6,370,767 data points in total.

\subsection{Variables}
Along with the variables used in Experiment 1, we use the number of dependencies completed at each region.\footnotemark{} This value should predict the anti-locality effect \cite{konieczny-2000-locality,konieczny-doring-2003-anticipation,vasishth-lewis-2006-argument}, i.e., the processing facilitation due to preceding elements that are related to the current input. In fact, previous studies on Japanese naturalistic texts show robust facilitatory effects of dependency completion \cite{asahara-etal-2019-bccwj,isono-etal-2025-modeling}. We predict that this facilitatory effect is weaker for participants who speed up in the face of a high maintenance cost, since these participants are sacrificing prediction.

\footnotetext{We count dependencies completed at each UD token and then take the maximal value for each region. Thus we avoid counting most completions of region-internal dependencies.}

\subsection{Analysis}
We first identify whether each participant slows down or speeds up when faced with structures that impose a high maintenance cost. We model each participant's reading time separately using two regression models, one containing the number of predicted heads in addition to control variables, and another containing the number of incomplete dependencies in addition to control variables.\footnotemark{} Since we are interested in the sign of the coefficients, we split data from each participant into 10 folds, fit models for each fold separately, and conduct permutation tests testing whether the mean of the coefficients is significantly above or below zero. This gives a typology of participants: for each of the two maintenance cost variables, each participant faced with a high maintenance cost slows down significantly, speeds up significantly, or shows no significant evidence of either.\footnotemark{}

\addtocounter{footnote}{-1}
\footnotetext{We fit two separate models for each type of maintenance cost since putting them together in one model can destabilize the coefficient estimates by the high correlation between them.}
\addtocounter{footnote}{1}
\footnotetext{We use a categorial typology rather than coefficients themselves since the absolute value of the coefficients can reflect the participant's general sensitivity to linguistic features.}

We then investigate whether this typology affects prediction. An estimate of the anti-locality effect for each participant is obtained by fitting another model that contains all the target and control variables plus the number of dependency completions. We then fit a linear model that predicts this anti-locality effect by each of the typology for the two maintenance cost variables.

\subsection{Results}
\begin{figure}[t]
  \includegraphics[width=\columnwidth]{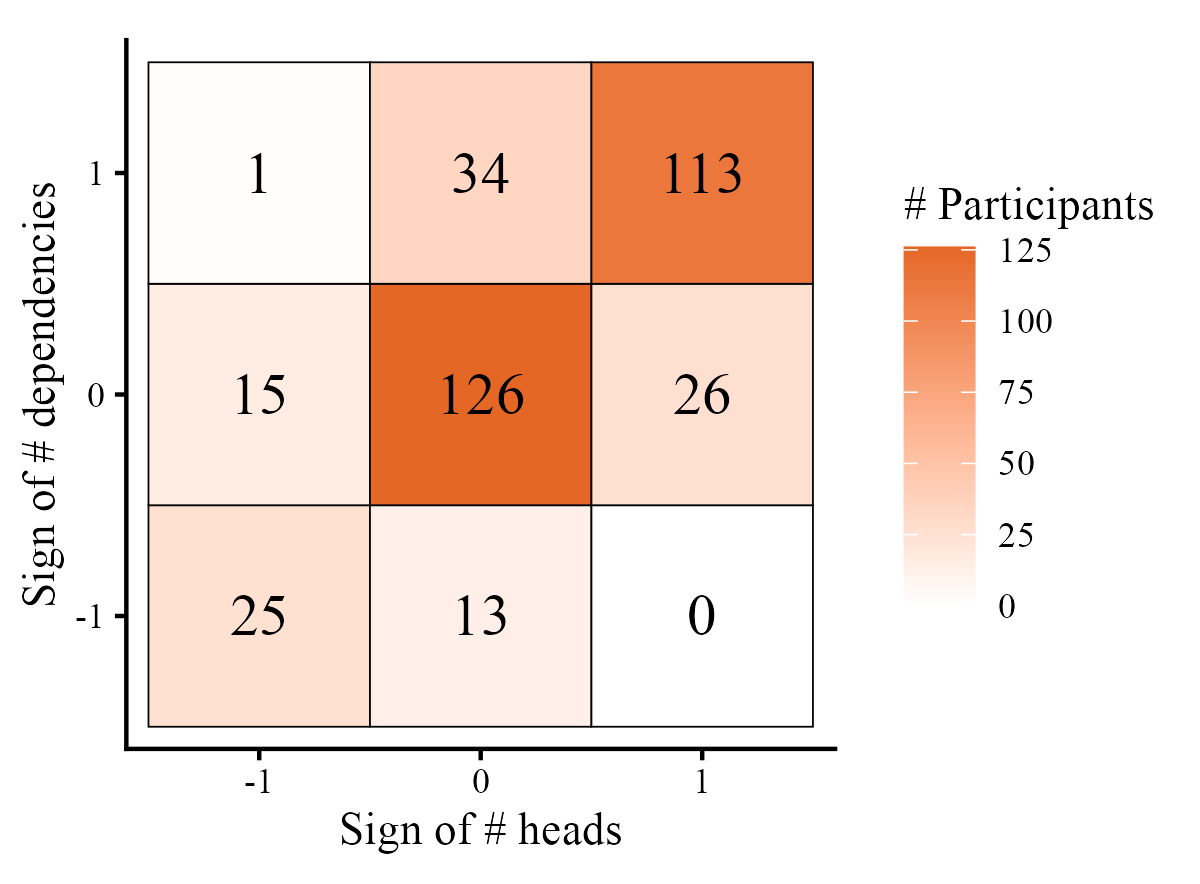}
  \caption{Classification of participants by their response to high maintenance cost. $1$ indicates significant slowdown, $-1$ indicates significant speedup, and $0$ indicates significance in neither direction.}
  \label{fig:typology}
\end{figure}

The typology of participants is summarized in Figure \ref{fig:typology}. When the number of predicted heads is high, 139 participants (39.4\%) show a significant slowdown while 41 (11.6\%) show a significant speedup. The trend is similar when analyzed in terms of the number of incomplete dependencies.

\begin{figure}[t]
  \includegraphics[width=\columnwidth]{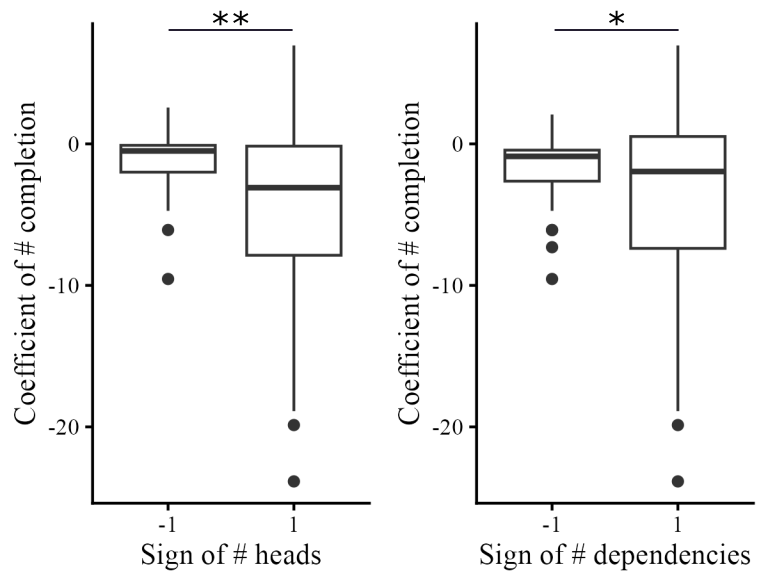}
  \caption{Distribution of anti-locality effect (y-axis) by participants' maintenance strategy (x-axis). The stars indicate the significance of the permutation test: ** for $p<0.01$, * for $p<0.05$}
  \label{fig:completion}
\end{figure}

The relation between this typology and the anti-locality effect is summarized in Figure \ref{fig:completion}. As expected, participants who slow down in the face of a high maintenance cost show a significantly stronger anti-locality effect, both when the typology is in terms of predicted heads ($p=0.001$) and incomplete dependencies ($p=0.049$).

\section{Experiment 3} \label{sec:experiment-3}
So far we have analyzed a Japanese dataset, but a natural question to ask is whether the observed patterns generalize to typologically different languages, as reviewers pointed out. We therefore additionally analyzed the Natural Stories dataset \cite{futrell-etal-2021-natural}, which contains self-paced reading time by 181 native speakers of English reading 10 stories (10,245 words). The methods are the same as Experiments 1 and 2 unless otherwise specified. The figures are in the Appendix.

\subsection{Variables}
\paragraph{Maintenance cost}
English often uses adjuncts that adjoin from the right, unlike in Japanese. We do not count those right adjuncts as predicted or forming an incomplete dependency.\footnotemark{} This is because right adjuncts are generally unpredictable. For example, in \textit{the professor read the book carefully}, the adverbial \textit{carefully} is optional. It is not sensible to assume that \textit{read} triggers different number of predictions depending on whether an optional adjunct follows in the actual continuation.

\footnotetext{The following dependency types are regarded as adjunctions: \texttt{appos, acl, acl:relcl, advcl, advmod, amod, cc, compound, compound:prt, conj, dep, det:predet, discourse, nmod, nmod:npmod, nmod:poss, nmod:tmod, nummod, parataxis, punct}.}

\paragraph{Control variables}
The following control variables are used: the position of the region in the story and in the sentence, the number of characters in the region, the negative log frequency of the word, and GPT-2 surprisal of the word.

\subsection{Results}
In the initial analysis, neither the number of predicted heads or the number of incomplete dependencies showed significant predictive power. We then limited the analysis to content words,\footnotemark{} since the heavy use of functional \emph{words} is a major difference from Japanese, which mainly use functional \emph{morphemes}. The results reported below are therefore exploratory.

\footnotetext{Words of the following POS are included (\texttt{x} stands for any suffix): \texttt{CD, JJx, NNx, NP, RBx, VBx}.}

\begin{figure}[h]
  \includegraphics[width=\columnwidth]{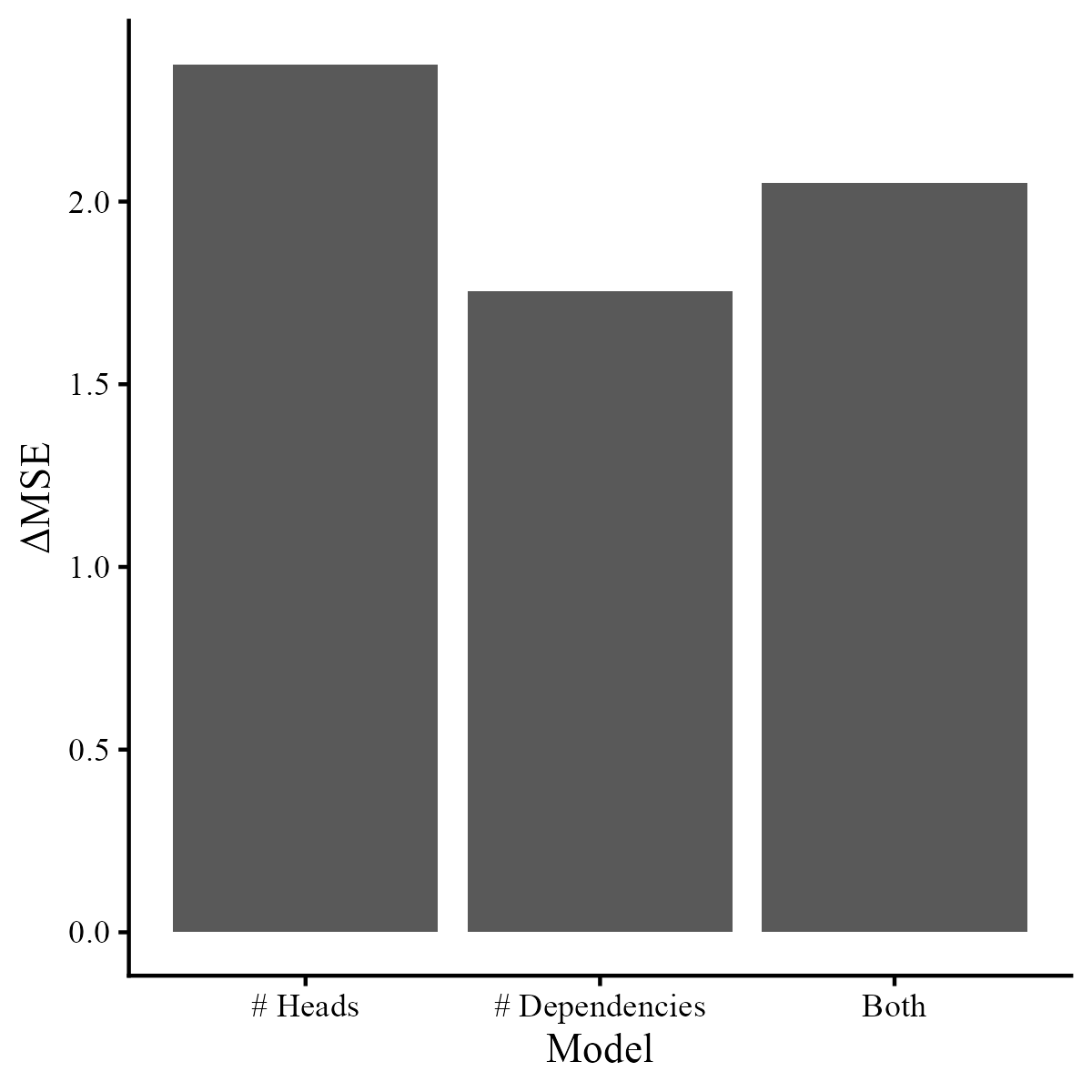}
  \caption{$\Delta$MSE per word by model (English). None of the comparisons were significant.}
  \label{fig:dmses.en}
\end{figure}

\paragraph{Psychometric predictive power}
The number of predicted heads shows a significantly positive $\Delta$MSE ($p=0.03$), while the number of incomplete dependencies is only marginally significant ($p=0.07$). The difference is not significant. Adding one to the other does not significantly improve the model fit (Figure \ref{fig:dmses.en} and Table \ref{tab:dmses.en}).

\begin{figure}[h]
  \includegraphics[width=\columnwidth]{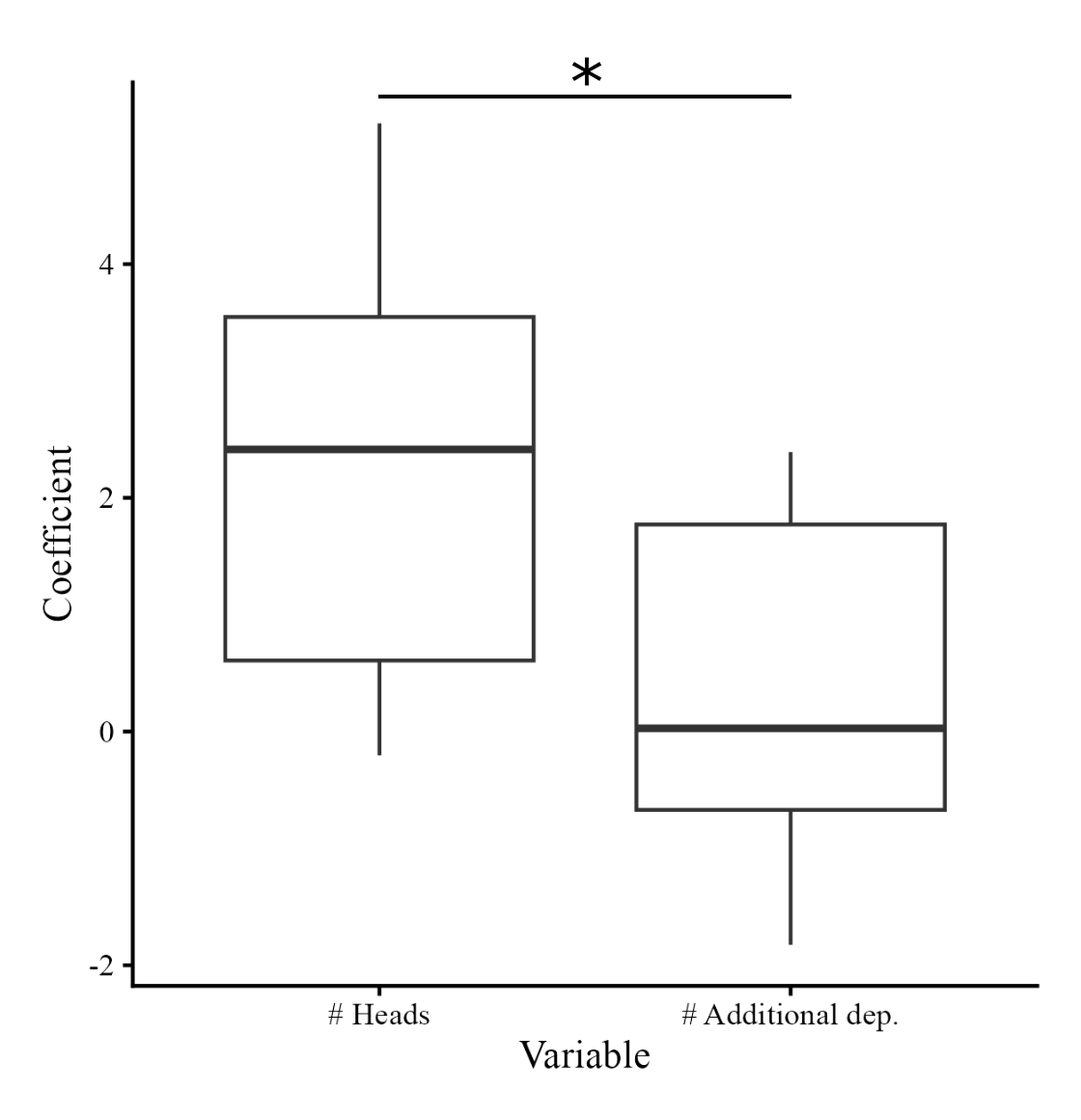}
  \caption{Distribution of estimated coefficients of the metrics of maintenance cost (English). The star indicates the significance of permutation test: * for $p<0.05$}
  \label{fig:coefficients.en}
\end{figure}

\paragraph{Coefficients}
The number of predicted heads shows a consistently positive effect, while the number of incomplete dependencies does not. The difference is significant ($p=0.01$) (Figure \ref{fig:coefficients.en}).

\begin{figure}[h]
  \includegraphics[width=\columnwidth]{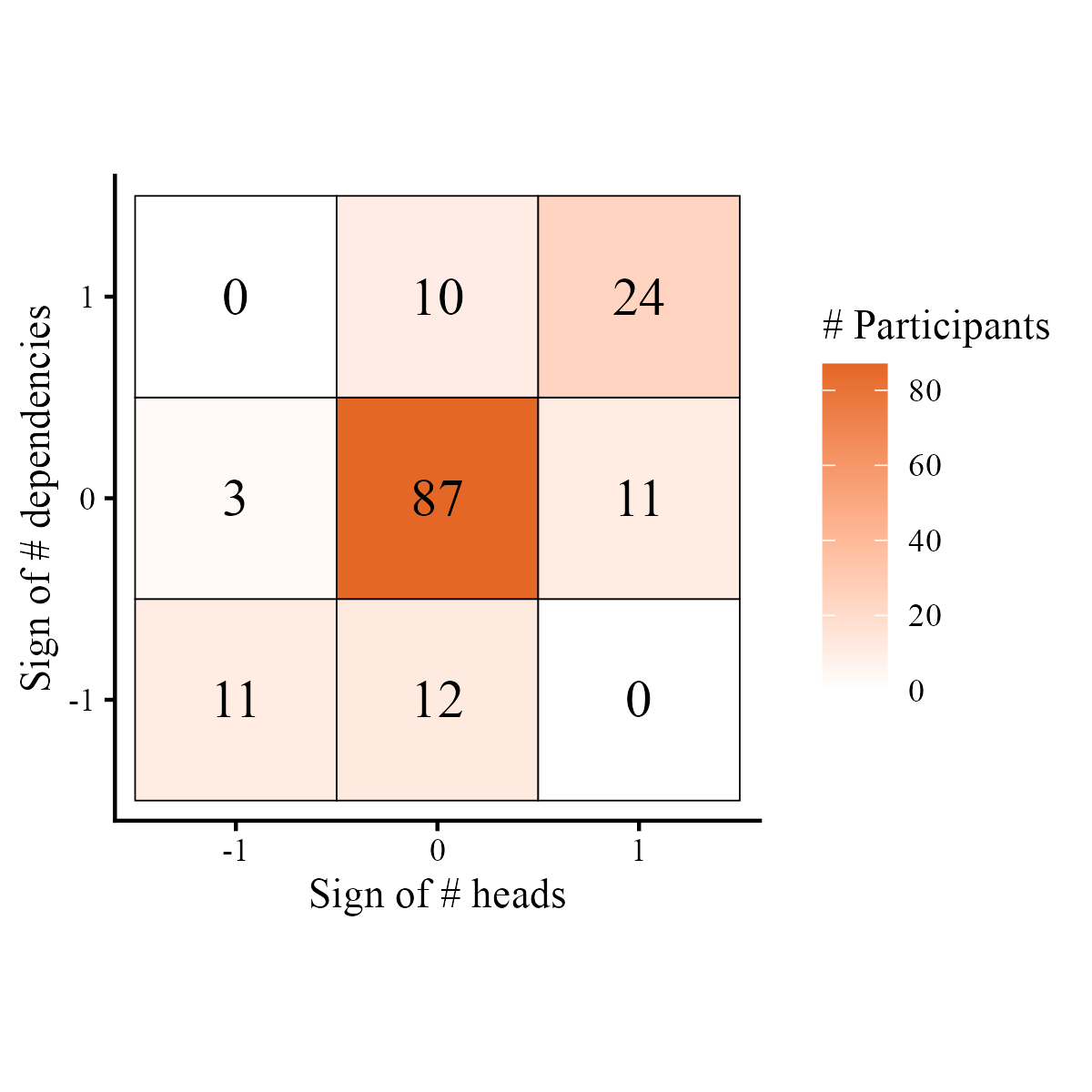}
  \caption{Classification of participants by their response to high maintenance cost (English). $1$ indicates significant slowdown, $-1$ indicates significant speedup, and $0$ indicates significance in neither direction.}
  \label{fig:typology.en}
\end{figure}

\paragraph{Variation between participants.}
22.2\% of participants consistently slow down when the number of predicted head is high, while 8.9\% speed up (Figure \ref{fig:typology.en}). The tradeoff between the antilocality effect and the slowdown due to high maintenance cost is not observed: the coefficient for the antilocality effect is numerically higher for participants who slow down. The difference is marginally significant when the number of dependencies is analyzed ($p=0.06$), but not significant when the number of predicted heads is analyzed.

In summary, the significant effect of the number of predicted heads is consistent with the view that syntax helps memory-efficient processing. Beyond that, we do not observe the interesting results observed in Japanese. We consider this as pointing to issues to be resolved to generalize the current theory to languages like English (see Limitations).

\section{Discussion} \label{sec:discussion}
We showed that the reading slowdown due to information maintenance in a head-final language is affected by both the number of predicted heads and incomplete dependencies. The number of predicted heads is associated with a larger impact on reading times, though the two factors cannot be reduced to the other. We further showed that readers have different strategies for maintenance, and their choice affects how much they benefit from contextual prediction. These results are consistent with the view that the information maintenance of a rational reader can be modeled by a syntactically-informed strategy aiming at better prediction or integration of future inputs.

\subsection{Syntax is a good guide}
We analyzed syntax-based metrics of maintenance cost. To be clear, we are not strongly committed to the mental reality of syntactic structure \cite[cf.][]{chomsky-1965-aspects}. Rather, our point is that, if syntactic structures exist in language at least as ``real patterns'' \cite{futrell-mahowald-2025-linguistics}, the maintenance strategy of a rational reader should closely align with it.

It is interesting in this regard that the hierarchical syntactic structure by design offers an efficient maintenance strategy. A crucial feature of syntax exploited here is that only the top element of a dependency (sub)structure can in principle interact with the outside. If human language was less restrictive, freely allowing crossing and circular dependencies, the processor would have to maintain every word it has recognized, and would be quickly overburdened. Cognitive limitations on information maintenance can thus provide an explanation of why language has hierarchical syntax \cite{christiansen-chater-2016-now}.

\subsection{Related work} \label{sec:related}
The current study shares with the resource-rational lossy-context surprisal (RR-LCS) model \citep{hahn-etal-2022-resource} the view that human language users make a rational use of memory to minimize prediction error \cite[also see][]{xu-futrell-2026}. We believe that the current work is complementary to that line of work. Like ours, the RR-LCS model assumes that the processor minimizes prediction error (surprisal) by retaining important words in the context. The key difference is that the RR-LCS model implements this strategy using machine learning techniques. The prediction error is operationalized as GPT-2 surprisal, and the retention probability for each word is optimized using a dedicated neural network. Admittedly, a machine learning approach may give a more precise estimation of the rational retention strategy than a syntax-based approach, taking lexical properties of the words into account.
% But it remains a computational-level theory \cite{marr-1982-vision}.
% The syntax-based approach is a step toward an algorithmic-level theory of a resource-rational processing.
An advantage of the syntax-based approach proposed here is that it can be transparently embedded in an algorithmic-level theory of resource-rational processing \cite{marr-1982-vision}. It can also be connected to theoretical investigations about why language has structure that it has, as argued above.

The current results show that both predicted heads and incomplete dependencies slow processing. This is in a direct opposition to the observation of \citet{nakatani-gibson-2008-distinguishing}. In a controlled experiment in Japanese, they observed that additional incomplete dependencies result in \emph{faster} reading time. They used sentences like \Next (in the actual stimuli they are embedded within a matrix clause) and manipulated the presence of additional nouns preceding the verb, indicated in parentheses in \Next:

\exg.
(zimusyo-no) sinnyusyain-ga (zimusyo-de) (kokyaku-ni) tyumonsyo-o hassosita\\
(office-\textsc{gen}) freshman-\textsc{nom} (office-at) (client-\textsc{dat}) order.sheet-\textsc{acc} sent\\
`the freshman (in the office) sent the ordersheet (to the client) (in the office)'

They observed that the accusative noun (\textit{tyuumonsyo-o}) was read faster when a locative or dative noun precedes it. They take this result as evidence against incomplete dependencies as a determinant of maintenance cost.

The faster reading time could possibly be a predictability effect, however. The presence of a dative noun, for example, reduces the probability of another dative noun to follow. This can result in an increase in the relative probability of an accusative noun, making an accusative noun more predictable than when a dative noun does not precede \cite[see][for a similar argument regarding English PPs]{levy-2008-expectation}. Such a facilitation effect can mask the underlying maintenance cost.\footnotemark{} Unlike their study, the current study controls for predictability by including GPT-2 surprisal in the regression model. The significantly positive coefficient found in the current analysis suggests that incomplete dependencies do impose maintenance cost. A factorial experiment that controls for predictability would be supplementary to the current corpus-based work in understanding the contribution of the two factors of maintenance cost.

\footnotetext{This remains a speculation since the raw data is not publicly available.}

\section{Conclusion}
Faced with fleeting language, a rational language user should use their limited memory wisely. We hypothesized that syntax offers a memory-efficient strategy of memory use, and our results indeed suggest that many readers of a head-final language can be modeled as adopting such a strategy.

\section*{Limitations}\label{sec:limitations}
The patterns we observed in Japanese (independent effects of the two types of maintenance cost; tradeoff between antilocality and slowdown for maintenance) were not replicated in English. We do not take this result as immediately undermining the notion of syntactically-guided memory-efficient processing, but as highlighting some issues in implementing the idea in typologically diverse languages. One obstacle for a non-head-final language like English is the argument/adjunct distinction. When a head requires an argument to follow, it counts as a predicted head; when an adjunct follows a head, it need not be predicted. We relied on the UD annotation to detect right adjuncts and excluded them from predicted heads, but the distinction is not clear-cut~\citep{macdonald_lexical_1994,manning-2003}, causing a problem for estimating maintenance cost. Another difference that might account for the different results with Japanese and English is that functional morphemes are suffixes in the former but words in the latter. In the current formulation, only the latter counts as triggering prediction. But functional words carry relatively little information about the likely continuation compared to content words (e.g., \textit{the} versus \textit{delicious}). The English results may particularly be affected by the fact that the current approach does not take such differences into account.

Another problem, common to both Japanese and English, is that the we only considered the gold parse. If a reader holds multiple parses at a time, with a probability assigned to each of them, there is no obvious way to calculate the aggregate maintenance cost for all the parses. Weighting the maintenance cost for each parse by its probability is not necessarily appropriate, since there is no a priori reason to assume that a parse with lower probability occupies less memory space. How to calculate maintenance cost under parallel parses is a topic for future work.

These problems can be addressed if we give up a syntax-based theory of storage and move to an information-theoretic one, using a large language model to estimate predictive information. Our attempt in this direction has already proved to be promising, but the number of predicted heads remained independently significant \citep{kajikawa-etal-2026}. Future work should probably integrate syntactic and information-theoretic perspectives for a deeper understanding of memory-efficient processing in humans.

An alternative account of storage cost we did not pursue here is that holding multiple heads or dependencies of the same kind is costly \cite{stabler-1994-finite, lewis-1996-interference}. Although the intuition of this idea is partially covered by counting the number of predicted heads, it is beyond the scope of the current study to develop a specific definition of the same kind of heads or dependencies that can be tested against naturalistic data.

We focused on self-paced reading data in this study since the notion of storage is most straightforward in a paradigm that does not allow regression. In the eye-tracking-during-reading paradigm, readers may put less effort in storage since they can freely look back on the past context. A comparison with eye-tracking data is an interesting venue for future study.

\section*{Acknowledgments}
This study is dedicated to the late Akiyo Fukatsu, a cherished and warm-hearted colleague whose insatiable curiosity about language will continue to inspire and drive us.

We would like to thank the reviewers for their insightful suggestions. This work is supported by JSPS KAKENHI Grant number JP25K22996.

% Bibliography entries for the entire Anthology, followed by custom entries
%\bibliography{anthology,custom}
% Custom bibliography entries only
\bibliography{custom}

\section*{Appendix: Figures and Table for Experiment 3}
\begin{figure}[h]
  \includegraphics[width=\columnwidth]{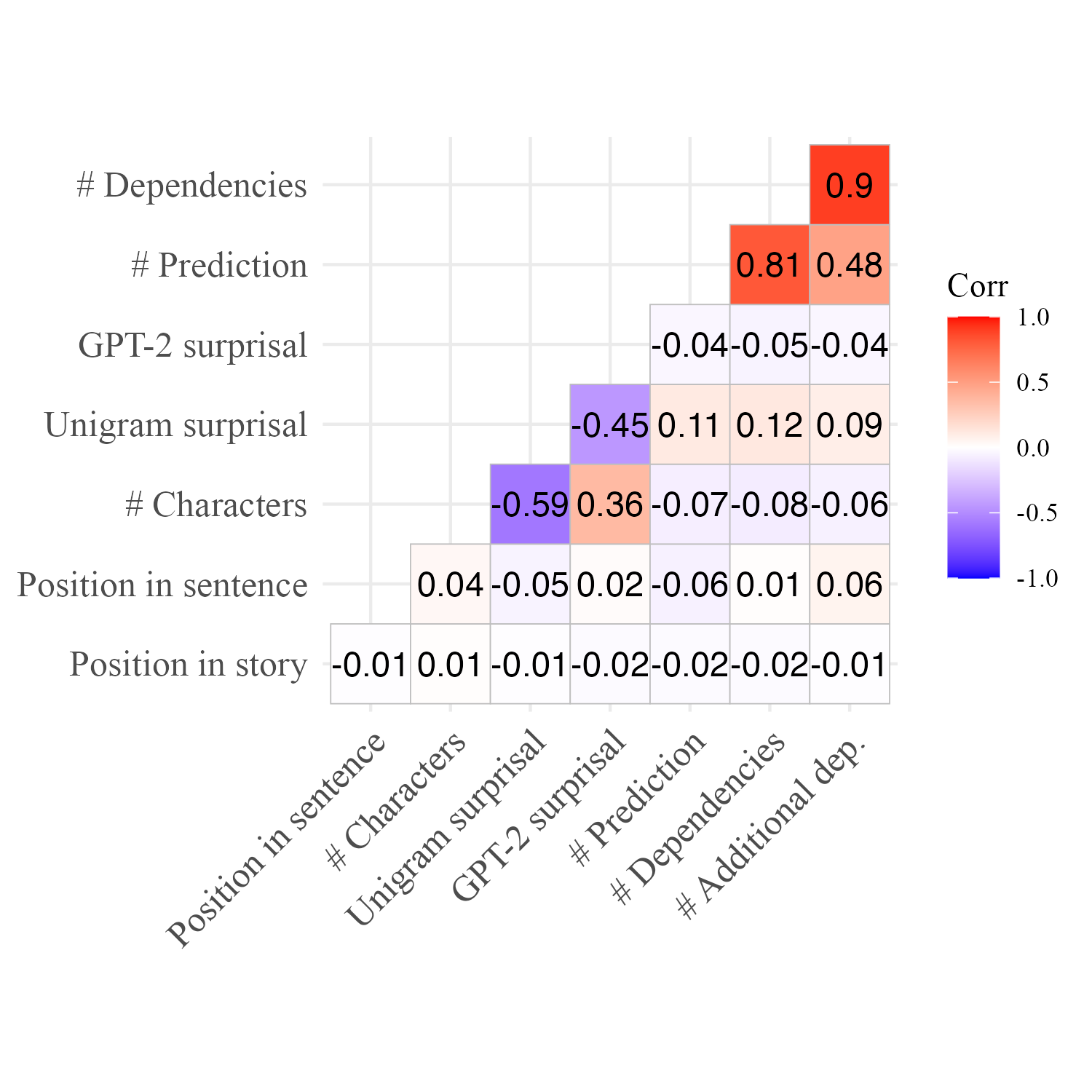}
  \caption{Correlations between variables used in the regression analysis (English). \# = number of.}
  \label{fig:correlations.en}
\end{figure}

\begin{table}[h]
    \centering
    \begin{tabular}{lr}
         & $\Delta$MSE \\
         \hline
         Position in the story & 290.1\\
         Position in the sentence & 0.1\\
         \# Characters & 61.4\\
         Unigram surprisal & 32.8\\
         GPT-2 surprisal & 74.0\\
         \hline
         \# Heads & 2.4\\
         \# Dependencies & 1.8\\
         Both & 2.1\\
         \hline
    \end{tabular}
    \caption{$\Delta$MSE of predictors.}\label{tab:dmses.en}
\end{table}

\begin{figure}[h]
  \includegraphics[width=\columnwidth]{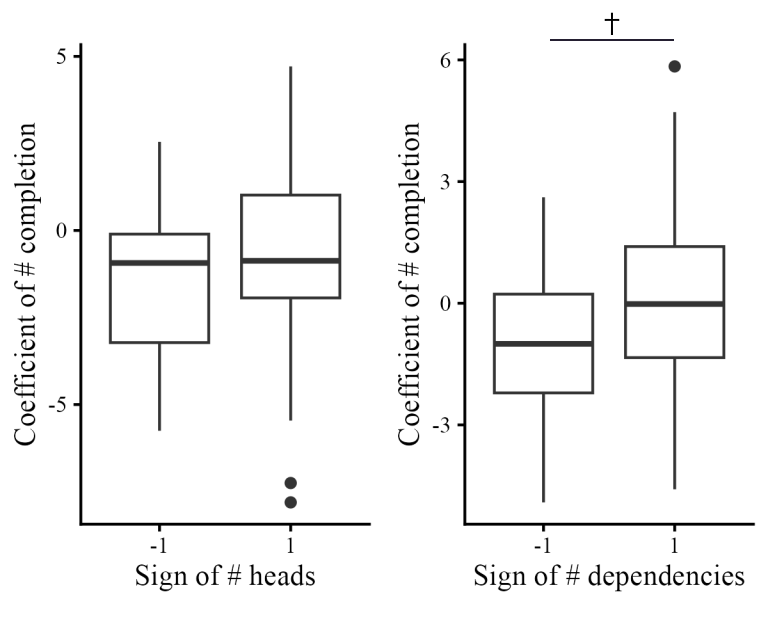}
  \caption{Distribution of anti-locality effect (y-axis) by participants' maintenance strategy (x-axis) (English). The star indicates the significance of the permutation test: $\dagger$ for $p<0.1$}
  \label{fig:completion.en}
\end{figure}

\end{document}